\documentclass[11pt,a4paper]{article}
\usepackage[hyperref]{acl2021}

\usepackage{times}
\usepackage{times}
\usepackage{color}
\usepackage{latexsym}

\usepackage{CJKutf8}
\usepackage{stfloats}
\usepackage{graphicx}
\usepackage{multicol}
\graphicspath{ {acl-ijcnlp2021-templates/images/} }
\usepackage[table,xcdraw]{}
\usepackage{microtype}
\usepackage{booktabs}
\usepackage{soul}
\usepackage{bbding}
\usepackage{ulem}
\usepackage{amsmath}
\usepackage{amssymb}
\usepackage{pifont}
\newcommand{\xmark}{\ding{54}}%
\newcommand{\cmark}{\ding{51}}%
\usepackage{multirow} 
\usepackage{makecell}
\usepackage{stfloats}
\usepackage{stackengine}

\usepackage{multicol}
\usepackage{tabularx}
\usepackage{threeparttable}
\usepackage{hyperref}

\aclfinalcopy 
\def\aclpaperid{3885} 

\newcommand\halfcorrect{
    \begin{minipage}{1em}
        \begin{center}
            \includegraphics[width=0.75\linewidth]{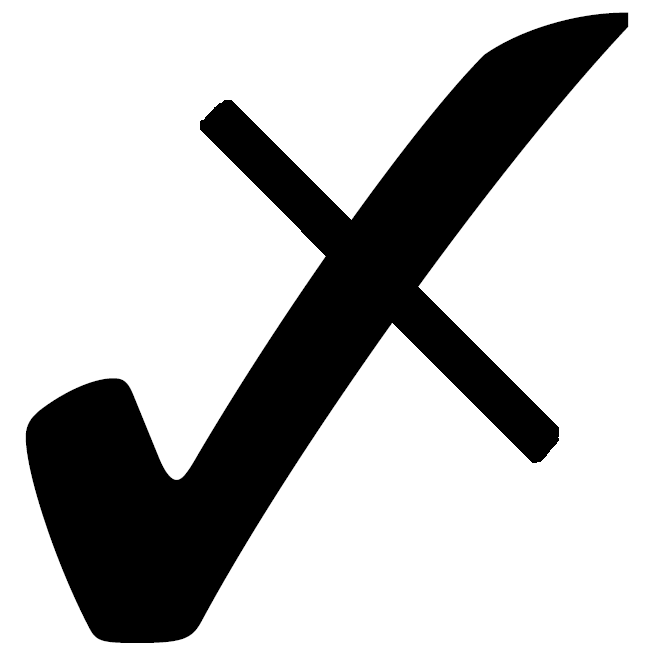}
        \end{center}
    \end{minipage}
}

\title{TGEA: An Error-Annotated Dataset and Benchmark Tasks for Text Generation from Pretrained Language Models}

\author{
  Jie He\textsuperscript{1}\thanks{\ \ Equal Contributions.}\ \ ,  
  Bo Peng\textsuperscript{2}\footnotemark[1]\ \ ,  
  Yi Liao\textsuperscript{2},  
  Qun Liu\textsuperscript{2},  
  \and Deyi Xiong\textsuperscript{1} \\  
  \textsuperscript{1}College of Intelligence and Computing, Tianjin University, Tianjin, China \\  
  \textsuperscript{2}Huawei Noah’s Ark Lab, Hong Kong, China \\  
  \texttt{\{jieh, dyxiong\}@tju.edu.cn} \\  
  \texttt{\{peng.bo2, liaoyi9, qun.liu\}@huawei.com}
}

\date{}

\begin{document}
\begin{CJK}{UTF8}{gbsn}
\maketitle
\begin{abstract}
In order to deeply understand the capability of pretrained language models in text generation and  conduct a diagnostic evaluation, we propose TGEA\footnote{The dataset is available at https://download.mindspore.cn\\ /dataset/TGEA/.}, an error-annotated dataset with multiple benchmark tasks for text generation from pretrained language models (PLMs). We use carefully selected prompt  words to guide GPT-2 to generate candidate sentences, from which we select 47K for error annotation. Crowdsourced workers manually check each of these sentences and detect 12k erroneous sentences. We create an error taxonomy to cover 24 types of errors occurring in these erroneous sentences according to the nature of errors with respect to linguistics and knowledge (e.g., common sense). For each erroneous span in PLM-generated sentences, we also detect another span that is closely associated with it. Each error is hence manually labeled with comprehensive annotations, including the span of the error, the associated span, minimal correction to the error, the type of the error, and rationale behind the error.  Apart from the fully annotated dataset, we also present a detailed description of the data collection procedure, statistics and analysis of the dataset. This is the first dataset with comprehensive annotations for PLM-generated texts, which facilitates the diagnostic evaluation of PLM-based text generation. Furthermore, we use TGEA as a benchmark dataset and propose a series of automatic diagnosis tasks, including error detection, error type classification, associated span detection, error rationale generation, to further promote future study on the automatic error detection and correction on texts generated by pretrained language models. 
\end{abstract}

\section{Introduction}
Pretrained language models \citep{devlin-etal-2019-bert,roberta,t5,gpt3},  which are trained  on a huge amount of data via self-supervised learning, have made remarkable progress on both natural language understanding (NLU) \citep{wang-etal-2018-glue,NEURIPS2019_4496bf24,long-webber-2022-facilitating,long-etal-2024-multi} and natural language generation (NLG) \citep{liu-lapata-2019-text,Weng_Yu_Huang_Cheng_Luo_2020,cao-etal-2020-pretrained}. 

On several NLU datasets, PLM-based neural models have gradually achieved human-level performance  in terms of automatic evaluation metrics (e.g., accuracy, $\rm F_1$) \cite{he2020deberta,zhang2021retro}. In order to deeply understand and analyze the capability of PLMs on NLU, a variety of more challenging NLU datasets have been proposed \cite{warstadt-etal-2020-blimp,cui-etal-2020-mutual,jain-etal-2020-scirex,olmpics,long-etal-2020-shallow,long-etal-2020-ted}. These datasets can be used not only to obtain knowledge on how PLM-based models work and what they learn, but also to define new NLU tasks and to serve as a benchmark for future progress.  For example, evaluating and analyzing PLM-based models on learning document structures with a carefully created benchmark test suite \citep{chen-etal-2019-evaluation,he-etal-2022-evaluating}, helps to develop new methods to enhance the capability of these models on discourse modeling \citep{iter-etal-2020-pretraining,long2024leveraginghierarchicalprototypesverbalizer}. Knowing the weakness of current PLM-based models in commonsense reasoning \citep{DBLP:conf/aaai/ZhouZCH20} has inspired people to develop various reasoning datasets \citep{cui-etal-2020-mutual,zhang-etal-2020-reasoning}. 

On the other hand, state-of-the-art PLMs are able to generate texts that are even not distinguishable from human-written texts by human evaluators \cite{gpt2,gpt3}. This makes us curious about the capability of PLMs on text generation. Are they really reaching human-level performance on text generation? In contrast to the studies of PLMs on NLU, research on the capability of PLMs on NLG is quite limited, especially in dataset building and diagnostic evaluation of text generation errors. 

\begin{figure}
\setlength{\belowcaptionskip}{-0.4cm} 
\centering
\includegraphics[scale=0.45]{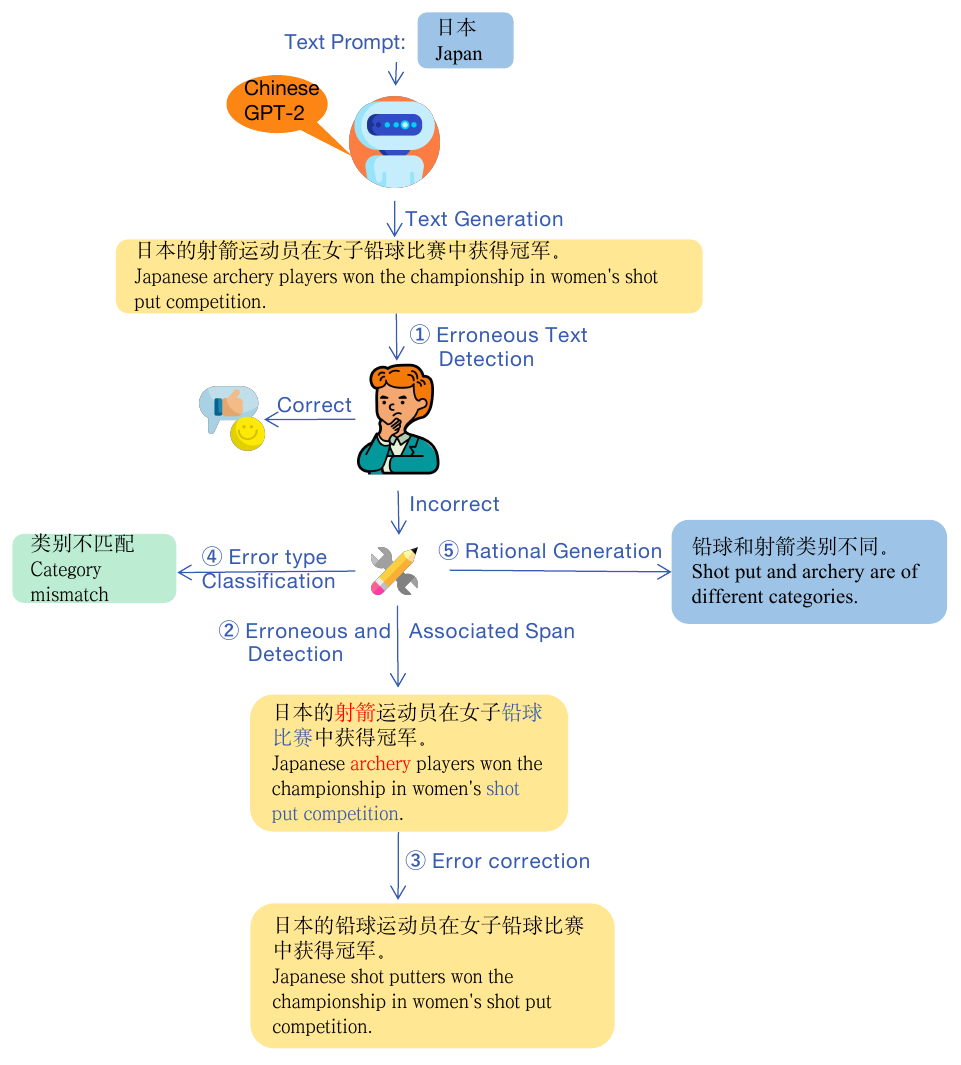}
\caption
{The different stages of the annotation process  for each machine-generated text according to the prompt in TGEA. Better viewed in color.
}
\label{figure1}
\end{figure}

In this paper, in order to recognize the perimeter of text generation capability of PLMs, we propose TGEA, an error-annotated dataset with multiple benchmark tasks for text generation from pretrained language models. The original raw data are collected from texts generated by a Chinese GPT-2 model. The entire data collection and annotation procedure is visualized in Figure \ref{figure1}.  The goals and contributions of building TGEA are as follows.

\begin{itemize}
\item
TGEA, to the best of our knowledge, is the first dataset built on machine-generated texts from state-of-the-art pretrained language models with rich annotations. The key interest of this dataset is detecting and annotating text generation errors from PLMs. Therefore it is different from conventional text generation datasets (e.g., Multi-News \citep{fabbri-etal-2019-multi}, TextCaps \citep{image-to-caption}) that are constructed to train models to learn text generation (e.g., generating texts from images or long documents). It is also different from grammatical error correction (GEC)  datasets  \cite{cgec,cweb} that are built from human-written texts usually by second language learners. 
\item
TGEA provides rich semantic information for text generation errors, including error types,  associated text spans, error corrections and rationals behind errors, as shown in Figure \ref{figure1}. Marking text spans that are closely related to erroneous words allows us to detect long-distance dependencies of errors or reasoning chains related to errors. Rationales behind errors directly explain why errors are annotated. All these error-centered manual annotations not only increase the interpretability of our dataset, but also facilitate a comprehensive diagnostic evaluation of pretrained language models on text generation. 
\item
We created an error taxonomy for TGEA, which covers 24 error types in a two-level hierarchy. With this error taxonomy, we not only obtain a high agreement on manual  error annotation but also recognize the strengths and weaknesses of GPT-2 on text generation by estimating a distribution over these 24 error types. Comparing our dataset with GEC datasets, we find that humans and GPT-2 have a very different error distribution, especially on errors related to commonsense reasoning.
\item
TGEA not only exhibits text generation errors from pretrained language models, but also can serve as a dataset to train various models to automatically detect and correct these errors, like GEC datasets for training models to automatically correct human errors. We define 5 benchmark tasks over our dataset, i.e., erroneous sentence detection, erroneous span and associated span detection, error type classification, error correction and error rationale generation. For all these tasks, we provide experimental results using state-of-the-art models as baselines.
\end{itemize}
\begin{table*}
\setlength{\belowcaptionskip}{-0.4cm}
\centering
\scriptsize
\begin{tabular}{l|ccccccc}
\bottomrule
Dataset& Task &\makecell{Commonsense\\Reasoning}&Rationales&\makecell{Machine-Generated \\Texts}& Domain &\#Sentences&Language\\
\hline
FCE&GEC&\xmark&\xmark&\xmark&Essay&34K&EN\\
AESW&GEC&\xmark&\xmark&\xmark&Journal articles&1.2M&EN\\
JFLEG&GEC&\xmark&\xmark&\xmark&TOFEL Exam&1,511&EN\\
CMEG&GEC&\xmark&\xmark&\xmark&Web doc/Essay&8K&EN\\
CWEB&GEC&\xmark&\xmark&\xmark&Web doc&13K&EN\\
CGEC&GEC&\xmark&\xmark&\xmark&Essay&0.71M&ZH\\
\hline
WSC&Coreference Resolution&\cmark&\xmark&\xmark&Open&273&EN\\
HellaSwag&Plausible Inference&\cmark&\xmark&\halfcorrect&WikiHow articles&70K&EN\\
Social IQA&Question Answering&\cmark&\xmark&\xmark&Social situations&38K&EN\\
CosmosQA&Reading comprehension&\cmark&\xmark&\xmark&Narratives&35K&EN\\
PIQA&Plausible Inference&\cmark&\xmark&\xmark&Physical situations&21K&EN\\
Abductive NLI&Plausible Inference&\cmark&\xmark&\xmark&ROCStories&200K&EN\\
WinoWhy&Reason Explanation&\cmark&\cmark&\halfcorrect&Open&2,865&EN\\
\hline
TGEA (ours)&Multiple tasks&\halfcorrect&\cmark&\cmark&Open&47K&ZH\\
\toprule

\end{tabular}
\caption
{\label{tab:related_works}Comparison between our dataset and other datasets.}
\label{table1}
\end{table*}
\section{Related Work}

Our work is related to GEC datasets in error annotation and correction (machine vs. human errors). It is also partially related to commonsense reasoning datasets that have been proposed recently in that our dataset includes commonsense reasoning errors and rationales behind these errors. Our dataset is not related to conventional text generation datasets \citep{neu-wiki,wiseman-etal-2017-challenges,parikh2020totto} for training text generation models. A comprehensive comparison to GEC datasets and commonsense reasoning datasets is shown in Table \ref{tab:related_works}.

\subsection{Grammatical Error Correction Datasets}
FCE \cite{fce} is an early large-scale English grammatical error correction dataset, where raw texts are produced by English learners taking the First Certificate in English exams. AESW \cite{aesw} is a GEC dataset from a professional editing company. In addition to common grammatical errors, AESW covers style issues  as it contains texts mainly from scholarly papers. JFLEG \cite{jfleg} is a GEC dataset built from TOFEL Exams, which does not force annotators to make minimal edits, preferring holistic fluency rewrites. CMEG \cite{cmeg} is different from  general grammatical error correction datasets with texts from second language learners. It uses articles or blogs (e.g., Wiki, Yahoo)) written by native English speakers to explore grammatical error phenomena in different domains. CWEB \cite{cweb} also uses website texts in English, such as blogs. The difference between CWEB and CMEG is that the percentage of erroneous tokens in the former is smaller than the latter as the purpose of CWEB is to study grammatical error correction in low error density domains. CGEC \cite{cgec} is a large-scale Chinese grammatical error correction dataset, derived from  wrong sentences written by Chinese learners in the process of learning Chinese  as a second language.

In addition to the difference in text sources (i.e., human-written vs. machine-generated),  other significant differences between our dataset and existing GEC datasets are that our dataset contains commonsense reasoning errors and provides associated text span annotations and rationales for errors, as shown in Table \ref{tab:related_works}.

\subsection{Commonsense Datasets}
A variety of commonsense datasets have been proposed.  \citet{roemmele_choice_2011} introduce COPA that focuses on commonsense causal reasoning. \citet{wsc} present Winograd Scheme Challenge (WSC), a dataset testing commonsense reasoning in the form of anaphora resolution. Winogrande, a larger version of WSC, is introduced by \citet{sakaguchi_winogrande_2020}, which  contains $\sim 44,000$ examples. Winowhy \citep{winowhy} asks annotators to provide reasons for their decisions to WSC. In this aspect, the differences of our dataset from Winowhy are twofold. First, we provide reasons for errors rather than correct decisions to anaphora. Second, we provide reasons for all text generation errors, rather than  only errors related to commonsense reasoning.

In addition to COPA and WSC-style datasets, many large crowdsourced datasets have been also proposed recently. CommonsenseQA \citep{talmor_commonsenseqa_2019}, a commonsense question answering dataset, has been constructed from ConceptNet. HellaSwag \citep{hellaswag} and Abductive NLI \citep{anli} evaluate commonsense reasoning in the form of natural language inference. CosmosQA \citep{cosmosqa} is a dataset with multi-choice questions that require commonsense reading comprehension. 

Beyond datasets for evaluating commonsense reasoning, there are other datasets providing commonsense knowledge. PIQA \citep{bisk_piqa_2020} focuses on physical commonsense knowledge while  SocialIQA \citep{siqa} on social commonsense knowledge.

Commonsense datasets in multiple languages or languages other than English have also been created  recently.  XCOPA \citep{xcopa} is a multilingual dataset for causal commonsense reasoning in 11 typologically different languages. Chinese commonsense datasets, such as Mandarinograd \cite{bernard_mandarinograd_2020} consisting  of 154 Chinese Winograd scheme examples and CLUEWSC2020 \citep{cluewsc} containing  1838 Winograd scheme  examples, have been proposed. 

In the aspect of commonsense reasoning, our dataset is different from the mentioned commonsense datasets in that we detect and annotate errors in machine-generated texts, which violates common sense, rather than creating examples to examine the commonsense reasoning ability of machines. 
\begin{table*}[]
\setlength{\belowcaptionskip}{-0.45cm}
\small
    \centering
    \begin{tabular}{|l|l|}
\bottomrule
Level-1 Error Type&Example \\
\hline
Inappropriate combination&医生当即将刘莉的\uwave{手术}[囊肿]\uline{切除}，并建议患者住院观察。\\
&The doctor \uline{removed} Liu Li's \uwave{surgery} [tumor] and suggested that the patient be hospitalized\\ 
&for observation.\\
\hline
Missing&在这里,有众多新闻记者和游客\uline{参加}\uwave{~~~}[活动]。\\
&Here, many journalists and tourists \uline{are taking part in} \uwave{~~~}[activities].\\
\hline
Redundancy&一些企业减员\uwave{增效}[]\uline{增效},使得企业利润增长了10\%以上。  \\
&Some enterprises have reduced staff and \uwave{increased efficiency}[] \uline{increased efficiency}, making \\
&their profits increase by more than 10\%.\\
\hline
Discourse Error&他说自己最喜欢安阳的乡间\uline{小路}，是最美的\uwave{山峦} [路]。\\
&He said that he likes the country \uline{roads} in Anyang best, and it is the most beautiful  \\
&\uwave{mountain} [road]. \\
\hline
Commonsense Error&在国际市场上，如果信用等级越\uwave{高} [低]，投资者在投资时就越\uline{不会太放心}。\\
&In the international market, the \uwave{higher} [lower] the credit rating, the \uline{less reassured} \\
&investors are.\\
\toprule
    \end{tabular}
    \caption{Examples of level-1 error types in TGEA. Underwaved words are erroneous words while underlined words are associated words. Words in ``[]" are corrections to erroneous words.}
    \label{table2}
 \end{table*}
\vspace{-1em}
\section{Dataset Creation}
\subsection{Error Taxonomy}
Before crowdsourced workers manually annotate errors in machine-generated texts, we need to create an error taxonomy for such error coding. Three principles are used to guide the design of the error taxonomy: coverage, exclusiveness and easiness. The coverage rule requires that the error system can cover almost all different types of errors in machine-generated texts. The exclusiveness requirement indicates that each error type is not overlapping with other error types in the taxonomy. The final easiness principle means that the error coding system is easy to be used by annotators. With these three principles and aid from a linguist, we created an error taxonomy in a two-level hierarchy, which was revised in our pre-annotation stage. 

The first level of the error taxonomy includes 5 error types.
\begin{itemize}
\item {\textit{Inappropriate combination.}} This type of errors suggests that two words/phrases are syntactically or lexically inappropriately combined in a sentence. Such errors include not only lexical collocation errors but also long-distance syntactic constituency combination errors (e.g., inappropriate subject-object combination). This error type is similar to ``replacing" error in some GEC datasets (e.g.,  CWEB \cite{cweb}) as one element of an inappropriate combination should be usually replaced with other expressions. As we want to find text spans associated with erroneous words/phrases, we term this error type as ``inappropriate combination". We further divide this error type into five subtypes at the second level.

\item {\textit{Missing.}} Grammatical constituencies or words are missing. 5 subtypes are defined under this error type.

\item {\textit{Redundancy.}}  Words or phrases are unnecessary. 5 subtypes are also defined.

\item {\textit{Discourse Error.}} This error type is defined for inter-sentential cohesion/coherence errors (e.g., coreference errors, incorrect discourse connectives).

\item {\textit{Commonsense Error.}}  This error code is for errors related to commonsense reasoning. We divide this error type into 8 subtypes according to the type of commonsense knowledge type required (e.g., time, spatial, number).
\end{itemize}

All other errors that cannot be categorized into the aforementioned error types are grouped into ``Other". Table \ref{table2} displays examples for the above defined error types.  24 error subtypes are displayed in Figure \ref{figure2} and examples of these subtypes are shown in Appendix.
\subsection{Machine-Generated Text Collection}

Raw texts in our dataset  are collected from  a pre-trained Chinese GPT-2 (NEZHA-Gen)\footnote{\href{https://github.com/huawei-noah/Pretrained-Language-Model/tree/master/NEZHA-Gen-TensorFlow}{github.com/huawei-noah/Pretrained-Language-Model/tree/master/NEZHA-Gen-TensorFlow}}, which generates texts according to a system prompt.  NEZHA-Gen has 12 layers and 12 attention heads  and is  trained on Chinese Wikipedia and news data (see Appendix for more  details on the hyperparameters of NEZHA-Gen). As it is easy for NEZHA-Gen to generate high-quality texts with high-frequency prompt words, we create a list of prompt words according to their frequency to guarantee that there are sufficient erroneous sentences in collected raw texts.  By doing so, we have found that GPT has a better chance to generate wrong sentences with such prompts. Specifically, we have randomly sampled 2M sentences from the data used to train NEZHA-Gen. The sampled sentences are then word-segmented and POS-tagged by Baidu LAC tool\footnote{\href{https://github.com/baidu/lac}{github.com/baidu/lac}} \citep{jiao2018LAC}. We then select and sort nouns in a descending order according to their frequencies in the sampled corpus.  Nouns ranking in the range of top [40\%, 60\%] are selected  as prompts.

We further filter out noisy texts from texts generated with these selected prompts. Noisy texts are either texts containing no more than 15 characters or texts where Chinese characters account for less 70\% of all characters.

\subsection{Error Annotation}
There are 5 stages in error annotation, as shown in Figure \ref{figure1}. We introduce each of them in this subsection. 

(1) \textbf{Erroneous text detection.} Texts generated by NEZHA-Gen with prompt words are present to annotators one by one. The first stage of annotation is hence to detect erroneous texts for subsequent annotations. Corresponding tags are annotated  for texts being manually checked.

(2) \textbf{Erroneous and associated span detection.} The next task for annotators is to detect erroneous and associated text spans in detected erroneous texts. For erroneous span detection, as a text may contain several spans that can be edited or the text can be corrected in different ways, which span should be regarded as erroneous is closely related to the way that we correct the text. Therefore, the basic principle that guides the annotation of erroneous spans is also the rule that we use for error correction: making minimal edits, which is also used in GEC datasets \cite{cweb,jfleg}. In addition to the minimal edit principle, we also provide the following specific rules for annotators:
\begin{itemize}
\setlength{\itemsep}{0pt}
\setlength{\parsep}{0pt}
\setlength{\parskip}{0pt}
\item If annotators feel that a text is ambiguous and that it is difficult to correct the text, the text can be discarded without any further annotations.
\item If there are several spans that can be edited, the first erroneous span is preferred to be edited. 
\item If the number of errors to be corrected in a text is larger than 4, the text is removed.
\end{itemize}
Following these rules, annotators have removed 4,291 texts, which account for only 8.36\% of all detected erroneous texts in the first stage. 

In addition to erroneous span annotation, unlike GEC datasets \citep{aesw,cgec}, we also detect a text span that is closely related to the already detected erroneous span with respect to the error, and term this span as ``associated span". In Table 2, we show examples with annotated erroneous and associated text spans. For an inappropriate combination, the associated span is usually a span that should not co-occur with the erroneous span.  

(3) \textbf{Error correction.} After detecting erroneous spans in a given text, annotators are required to make corrections following the minimal edit principle. Annotators are also required to use common words for error correction to make the corrected text as fluent as possible. 

(4) \textbf{Error type classification.} Once annotators detect both erroneous and associated spans as well as provide corrections, they are becoming quite aware of these errors. Hence, we now ask them to categorize the annotated errors into error types defined in our error taxonomy. First, they select the primary type from the level-1 error types. Then, if there are level-2 error subtypes, annotators continue to select a subtype. We observe that errors annotated
with ``other'' only account for 5.70\%, suggesting that our error taxonomy has  good coverage.

(5) \textbf{Rationale generation.} Partially inspired by previous datasets that provide explanations together with corresponding annotations, e.g.,  e-SNLI \citep{esnli}, Winowhy \citep{winowhy} and R4C \citep{r4c}, we ask annotators to give a reason for each error to justify their annotations. To the best of our knowledge, no GEC datasets provide explanations for error corrections. We believe that annotated rationales can be used to improve the interpretability of neural models trained on our dataset.

\begin{table}[]
\setlength{\belowcaptionskip}{-0.4cm}
\scriptsize
    \centering
    \begin{tabular}{|l|c|c|}
    \bottomrule
         Task&IAA (\%)&Kappa (\%)  \\
         \hline
         Erroneous text detection&87.5&62.1\\
         \multirow{2}{*}{\makecell{Erroneous and associated \\span detection}}&\multirow{2}{*}{51.2}&\multirow{2}{*}{--}\\
         &&\\
         Error type classification&73.3&55.7\\
    \toprule
    \end{tabular}
    \caption{Inter-annotator agreement results.}
    
\label{table3}
\end{table}
\subsection{Annotation Quality Control}
In order to ensure the quality of error annotations, we have adopted a very strict quality control protocol during annotation.  First, we train two reviewers with 1K machine-generated texts. The annotation consistency of the two reviewers on the 1K texts is very high, with an average IAA of 92.3\% and Cohen's Kappa \citep{kappa} of 82.6\%  across the annotation tasks (1), (2) and (4). For the texts annotated by the two reviewers, we have conducted an evaluation. The average accuracy of all tasks is 96.3\% and 97.4\% respectively. 

Second, 200 candidate workers participate in a pre-annotation stage. The two reviewers will review annotations from these participants to distinguish whether the annotation is correct or not. Only participants who have reached an accuracy of \textgreater 90\%  in every tasks can join in the next stage. As a result, 20 participants have passed the training in the pre-annotation stage. We then divide them into two groups and ask them to annotate the same 500 texts. The inter-annotator IAA and Cohen's Kappa are shown in Table \ref{table3}, which suggests that the 20 annotators are ready for final annotation. 

Third, in order to further ensure annotation quality, we have carried out iterative verification and amendment. The two reviewers will review each annotated text. If they found the annotation is wrong, the unqualified data will be returned for amendment until they are qualified. 

Following this strict quality control protocol, we complete the annotation on 47K selected machine-generated texts. We randomly sample 1K annotated texts. The average accuracy over the three tasks (i.e., (1), (2) and (4)) is 89.6\%, 88.5\%, 84.3\% respectively. 

\section{Dataset Analysis}
\subsection{Dataset Statistics}
\begin{table}[]
    \centering
    \scriptsize
    \begin{tabular}{l|ccc|c}
    \bottomrule
        &Train&Dev&Test&All  \\
        \hline
        \#text &37,646&4,706&4,706&47,058 \\
        w/ 0 error&27,906&3,488&3,488&34,882\\
        w/ 1 error&8,413&1,055&1,052&10,520\\
        w/ 2 error&1,169&141&149&1,459\\
        w/ 3 error&141&18&15&174\\
        w/ 4 error&17&4&2&23\\
        \hline
        Tokens&966,765&120,889&121,065&1,208,719\\
        Vocab&44,598&16,899&16,745&48,547\\
        Avg. tokens&25.68&25.69&25.73&25.68\\
        \hline
        Avg. t.err&2.92&3.09&2.95&2.94\\
        Avg. t.assoc&4.30&4.39&3.89&4.27\\
        Avg. d.e-a&6.99&7.29&7.10&7.03\\
        Avg. t.rationale&8.74&8.72&8.75&8.74\\
    \toprule
    
    \end{tabular}
    \caption{Data statistics of TGEA. Avg.t.err/Avg.t.assoc: the average number of tokens in erroneous/associated text spans. Avg.t.rationale: the average number of tokens in rationales. Avg.d.e-a: the average distance between a erroneous span and its associated span.}
    \label{table4}
\end{table}
\textbf{Overall statistics.} We reshuffle all annotated texts and divide them into the training/dev/test sets with a proportion of 8:1:1. As shown in Table \ref{table4}, the training set contains 27,096 correct texts and 9,740 erroneous texts. Both the development and test set contain 4,706 texts, among which 1,218 texts are erroneous. Not surprisingly, most erroneous texts contain only one error.

After Chinese word segmentation via Jieba\footnote{\href{https://github.com/fxsjy/jieba}{github.com/fxsjy/jieba}}, there are 1,208,719 tokens in total. On average, there are 25.68 tokens in each text. 

\noindent\textbf{Annotation statistics.} As shown in Table \ref{table4}, each erroneous text span contains  2.94 tokens while each associated span is composed of  4.27 tokens. The average distance from an erroneous text span to its associated span is 7.03 tokens, which is about 1/3 of the average text length. 

\begin{figure}
\setlength{\belowcaptionskip}{-0.4cm}
\centering
\includegraphics[scale=0.35]{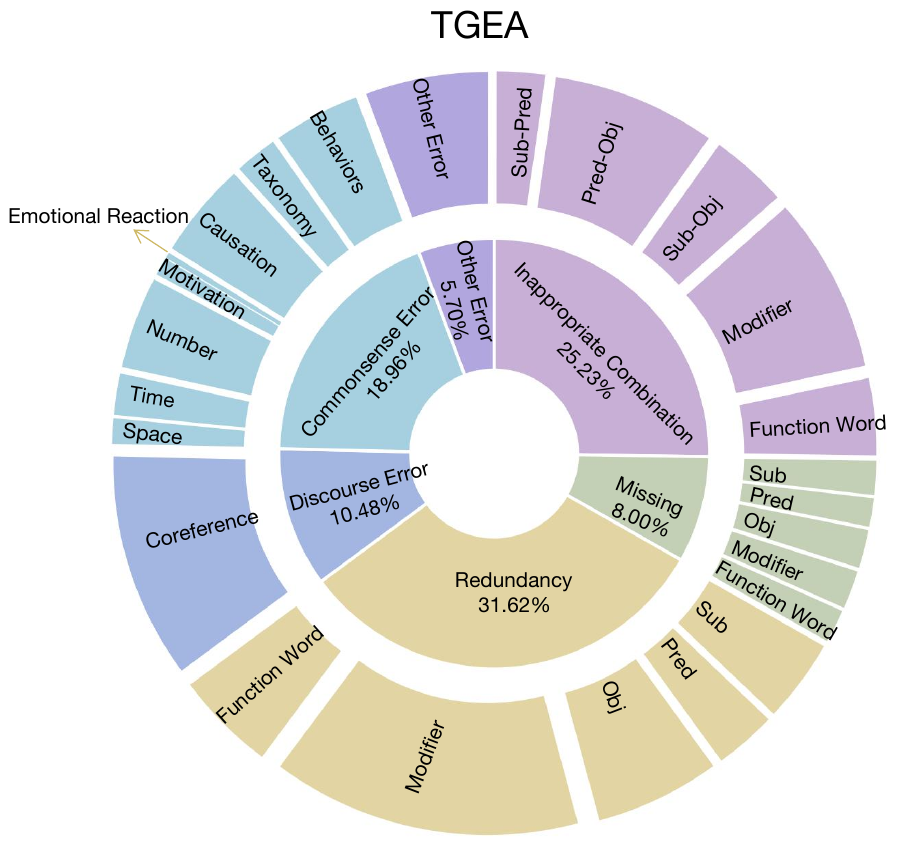}
\caption
{Distribution over the level-1 and level-2 error types  in TGEA.}
\label{figure2}
\end{figure}

\subsection{Error Type Distribution}
We further show the percentages of both level-1 and level-2 error types in Figure \ref{figure2}.  We observe that only 5.7\% cases cannot be categorized into our defined error types. The inappropriate combination, missing and redundancy error, which are the main error types in GEC datasets, account for 64.85\% in our dataset. In addition to these errors, we see 18.96\% commonsense errors and 10.48\% discourse errors, which are usually not very common in GEC datasets. However, these two types of errors with high percentages in our dataset suggest that pretrained language models can be further improved on both commonsense reasoning and discourse modeling. 

\section{TGEA as a Benchmark}
We use our dataset as a benchmark and propose 5 tasks that are defined for errors in texts generated by PLMs. We provide baseline results for these tasks in this section. 

We employ three BERT-style Chinese PLMs as baselines in our experiments, namely BERT-wwm-ext,
RoBERTa-wwm-ext-large developed by \citet{roberta-wwm-ext}
\footnote{\href{https://github.com/ymcui/Chinese-BERT-wwm}{github.com/ymcui/Chinese-BERT-wwm}} and ALBERT-Chinese-large\footnote{\href{https://huggingface.co/voidful/albert_chinese_large}{huggingface.co/voidful/albert\_chinese\_large}}.  For notational simiplicity, we denote them as $\rm BERT_{zh}$, $\rm RoBERTa_{zh}$ and $\rm ALBERT_{zh}$ respectively. Please refer to the Appendix for the model hyperparameter settings of each task. 

\subsection{Erroneous Text Detection}
\noindent\textbf{Task definition.} This is a text classification task to judge  whether a given text is erroneous. In order to avoid data imbalance, we use the same number of correct and erroneous texts for training.

\noindent\textbf{Model.}  The  three Chinese PLMs are used with standard text-classification fine-tuning.

\noindent\textbf{Results.} All models perform just \textless 14\% better than chance (random guessing), as shown in Table \ref{table5}.  We also provide human performance on this task. The best model $\rm RoBERTa_{zh}$  is worse than human performance by 26 points. This suggests that automatically detecting erroneous texts generated by pretrained language models is very challenging even in the balanced classification scenario.

\begin{table*}[]
\setlength{\belowcaptionskip}{-0.5cm} 

\small
    \centering
    \begin{tabular}{llcccccccccccc}
    \bottomrule
        \textbf{Task} &\textbf{Model} &\multicolumn{6}{c}{\textbf{Dev}}&\multicolumn{6}{c}{\textbf{Test}}\\
        \hline 
        &&\multicolumn{6}{c}{Accuracy (\%)}& \multicolumn{6}{c}{Accuracy (\%)}\\
        \cmidrule(r){3-8} \cmidrule(r){9-14}
        \multirow{4}{*}{\makecell{Erroneous\\text detection}}&Random&\multicolumn{6}{c}{50.00}&\multicolumn{6}{c}{50.00}\\
        &$\rm ALBERT_{zh}$&\multicolumn{6}{c}{63.59}&\multicolumn{6}{c}{63.30}\\
        &$\rm BERT_{zh}$&\multicolumn{6}{c}{65.15}&\multicolumn{6}{c}{64.94}\\
        &$\rm RoBERTa_{zh}$&\multicolumn{6}{c}{66.67}&\multicolumn{6}{c}{66.79}\\
        &\textbf{Human}&\multicolumn{6}{c}{92.35}&\multicolumn{6}{c}{93.57}\\
        \hline
        &&\multicolumn{3}{c}{\makecell{Erroneous\\class-$\rm F_1$ (\%)}}&\multicolumn{3}{c}{\makecell{Associated\\class-$\rm F_1$ (\%)}}&\multicolumn{3}{c}{\makecell{Erroneous\\class-$\rm F_1$ (\%)}}&\multicolumn{3}{c}{\makecell{Associated\\class-$\rm F_1$ (\%)}}\\
        \cmidrule(r){3-8} \cmidrule(r){9-14}
        \multirow{3}{*}{\makecell{Erroneous and\\associated\\span detection}}&Random&\multicolumn{3}{c}{01.71}&\multicolumn{3}{c}{04.23}&\multicolumn{3}{c}{01.74}&\multicolumn{3}{c}{04.22}\\
        &$\rm ALBERT_{zh}$&\multicolumn{3}{c}{27.36}&\multicolumn{3}{c}{27.44}&\multicolumn{3}{c}{28.10}&\multicolumn{3}{c}{26.24}\\
        &$\rm BERT_{zh}$&\multicolumn{3}{c}{27.85}&\multicolumn{3}{c}{26.93}&\multicolumn{3}{c}{27.66}&\multicolumn{3}{c}{25.30}\\
        &$\rm RoBERTa_{zh}$&\multicolumn{3}{c}{28.17}&\multicolumn{3}{c}{27.08}&\multicolumn{3}{c}{27.75}&\multicolumn{3}{c}{27.12}\\
        \hline
        &&\multicolumn{3}{c}{Accuracy (\%)}&\multicolumn{3}{c}{Macro-$\rm F_1$ (\%)}&\multicolumn{3}{c}{Accuracy (\%)}&\multicolumn{3}{c}{Macro-$\rm F_1$ (\%)}\\
         \cmidrule(r){3-8} \cmidrule(r){9-14}
        \multirow{3}{*}{\makecell{Error type\\classification}}&Random&\multicolumn{3}{c}{24.25}&\multicolumn{3}{c}{20.00}&\multicolumn{3}{c}{24.25}&\multicolumn{3}{c}{20.00}\\
        &$\rm ALBERT_{zh}$&\multicolumn{3}{c}{34.76}&\multicolumn{3}{c}{21.04}&\multicolumn{3}{c}{34.38}&\multicolumn{3}{c}{20.56}\\
        &$\rm BERT_{zh}$&\multicolumn{3}{c}{44.35}&\multicolumn{3}{c}{33.01}&\multicolumn{3}{c}{41.31}&\multicolumn{3}{c}{31.05}\\
        &$\rm RoBERTa_{zh}$&\multicolumn{3}{c}{44.44}&\multicolumn{3}{c}{36.10}&\multicolumn{3}{c}{44.16}&\multicolumn{3}{c}{37.20}\\
        \hline
        &&\multicolumn{2}{c}{P (\%)}&\multicolumn{2}{c}{R (\%)}&\multicolumn{2}{c}{$\rm F_{0.5}$ (\%)}&\multicolumn{2}{c}{P (\%)}&\multicolumn{2}{c}{R (\%)}&\multicolumn{2}{c}{$\rm F_{0.5}$ (\%)}\\
         \cmidrule(r){3-8} \cmidrule(r){9-14}
        \multirow{1}{*}{\makecell{Error correction}}&$\rm BERT_{zh}$\_GEC&\multicolumn{2}{c}{0.62}&\multicolumn{2}{c}{6.49}&\multicolumn{2}{c}{0.76}&\multicolumn{2}{c}{0.60}&\multicolumn{2}{c}{6.30}&\multicolumn{2}{c}{0.74}\\
        &$\rm RoBERTa_{zh}$\_GEC&\multicolumn{2}{c}{0.78}&\multicolumn{2}{c}{4.07}&\multicolumn{2}{c}{0.93}&\multicolumn{2}{c}{0.82}&\multicolumn{2}{c}{4.15}&\multicolumn{2}{c}{0.98}\\
        \hline
        &&\multicolumn{2}{c}{BLEU}&\multicolumn{2}{c}{Rouge-L}&\multicolumn{2}{c}{BERT\_Score}&\multicolumn{2}{c}{BLEU}&\multicolumn{2}{c}{Rouge-L}&\multicolumn{2}{c}{BERT\_Score}\\
         \cmidrule(r){3-8} \cmidrule(r){9-14}
        Rationale generation&NEZHA-Gen&\multicolumn{2}{c}{0.06\%}&\multicolumn{2}{c}{9.17\%}&\multicolumn{2}{c}{56.58\%}&\multicolumn{2}{c}{0.06\%}&\multicolumn{2}{c}{9.02\%}&\multicolumn{2}{c}{56.17\%}\\
    \toprule
    \end{tabular}
    \caption{\label{table5}Performance of benchmark models on the development and test set.}
\end{table*}

\subsection{Erroneous Span and Associated Span Detection}
\noindent\textbf{Task definition.} We define the detection of the two types of spans as a joint task as they are closely related to each other. The joint task is similar to named entity recognition (NER) (a sequence labeling task) and it requires to recognize the erroneous and associated text spans simultaneously. NER-style word-level tags are hence annotated for each erroneous text.

\noindent\textbf{Model.} The three Chinese PLMs with NER-like fine-tuning are evaluated for this task. Since this is a 3-class token classification task, we report class-$\rm F_1$ on erroneous and associated span. The class-$\rm F_1$ on class $X$ is calculated like a normal $\rm F_1$ for a binary classification task, by treating the target class $X$ as the positive class and all other classes as negative.

\noindent\textbf{Results.} As shown in Table \ref{table5}, all models are very poor in this task, indicating the difficulty of automatically detecting erroneous and associated spans. However, we have found that models can benefit much from the joint detection over the detection of a single type of span (either erroneous or associated span). Our preliminary experiments on the detection of only erroneous span show that the best model can only achieve 26.42\% erroneous class-$\rm F_1$ on the test set, while the joint task achieves 27.66\% erroneous class-$\rm F_1$ on the test set.
\subsection{Error Type Classification}
\noindent\textbf{Task definition.} Again this is a text classification task. We only perform classification over level-1 error types in the form of 5-way classification.

\noindent\textbf{Model.} We use models similar to the first task.

\noindent\textbf{Results.} The overall accuracy and Macro-$\rm F_1$ (shown in Table \ref{table5}) are very low. However, we find some error types are easier than others. The accuracy on the classification of redundancy errors is 53.91\%, the highest  among all error types. 

\subsection{Error Correction}
\textbf{Task definition.} This task is the same as GEC, which transforms an erroneous text into a correct sequence. 

\noindent\textbf{Model.} we use the state-of-the-art BERT-GEC model \citep{bert-gec}  as the baseline for this task, which is an encoder-decoder model using representations learned by PLMs as additional inputs. Following \citet{chinese-bert-gec}，we feed representations learned by $\rm BERT_{zh}$ and $\rm RoBERTa_{zh}$ into the BERT-GEC model.

\noindent\textbf{Results.} We report  precision, recall and $\rm F_{0.5}$ scores using the official Max-Match tool \citep{dahlmeier-ng-2012-better}. As shown in Table \ref{table5}, the best $\rm RoBERTa_{zh}\_GEC$ model achieves a very low $\rm F_{0.5}$ of 0.93\% and 0.98\% on the development and test set respectively. We speculate that the reasons for this are twofold. First, comparing with GEC data on human-written texts, our dataset is relatively small. Second, our dataset contains error types that are very different from those in previous GEC datasets \citep{cgec,cweb}. Punctuation, spelling and other word-character-level errors, which are easy to be corrected, are rare in TGEA although they are quite common in GEC datasets. In contrast, TGEA contains more complicated errors that can only be corrected with knowledge of common sense, long-distance or inter-sentential dependencies, etc. 

\subsection{Rationale Generation}
\noindent\textbf{Task definition.} This is a text generation task that directly generate an explanation with respect to text generation errors from an erroneous text. 

\noindent\textbf{Model.} We use NEZHA-Gen as the baseline for this task. We restructure annotated texts in our dataset in the form of $\{T,\allowbreak \text{这句话错误的原因是：},\allowbreak R\}$ ($\{T,\allowbreak \text{The reason behind the errors in this sentence is:},\allowbreak R\}$), where $T$ is an erroneous sentence, while $R$ is the error rational provided by annotators. We then fine-tune NEZHA-Gen on the reformatted training set and evaluate the fine-tuned model on the reformatted development and test set. We report BLEU \citep{papineni2002bleu}, Rouge-L \citep{lin2004rouge} and BERT\_Score \citep{bert-score}.

\noindent\textbf{Results.} It can be expected that results in these metrics will be very low due to the high difficulty of this task. We analyze generated texts from the baseline and find that generated rationales are usually much longer than reference rationales provided by human annotators. This could result in the low BLEU score since long hypotheses are penalized in BLEU computation. We also experiment zero-shot generation on the test set. The results are $\{\text{BLEU}=0.04\%, \text{Rouge-L}=6.83\%, \text{BERT\_Score}=54.27\%\}$, indicating that fine-tuning on the annotated training set can improve this task. We suggest that this generation task could be reformulated as a multi-choice question answering task by providing alternative rationales as distractors, similar to VCR \citep{zellers2019recognition}. We leave this to our future work.

\section{Discussion}

Since we use machine-generated texts for error annotation, hyperparameters of models (e.g., sampling strategies, model size), model types (e.g., GPT-2, GPT-3 or other PLMs for  text  generation), and genres of texts used to train PLMs, etc., all have impacts on generated texts and hence on error types and error distribution.

A straightforward way to mitigate this issue  is to collect raw texts from multiple models with different hyperparameters, neural architectures and text genres. This will lead  to an expanded dataset with a much larger number of instances to be manually annotated, which is expensive and time-consuming. Yet another issue with this is that it may result in a bunch of data due to inconsistency across different models and difficulty in setting the proportion of each data source.  

Instead, we focus on consistently annotating errors for texts generated from a single source. In order to make TGEA as general and representative as possible, we use GPT-2 that is not only currently state of the art in text generation but also easily available. We also adopt standard and widely-used hyperparameters (see Appendix for more details) for NEZHA-Gen to generate texts.

Additionally, we use a random sampling strategy with top $k=30$. For setting $k$, we have analyzed 500 examples with different values of $k$, and found that adjusting $k$ has a reasonable impact on the percentage of redundancy errors. Except for the extreme case of $k=1$, the types of errors and the distribution of them do not change significantly. Take commonsense errors as an example, which is the biggest difference from human-written texts. When $k$ varies in a range of  \{5, 10, 20, 30, 50\}, the percentage of commonsense errors is 18.6\% ± 5.8\%. Redundancy errors account for \textgreater95\% when $k=1$ (while commonsense errors account for 0.8\%), but sharply drop to 37.4\% as $k = 5$, and the form of repetition changes from same-word repetition to a mixed repetition of ``synonymous/same-word", suggesting that a simple repetition penalty may not be sufficient to deal with semantic redundancy. When $k \in \{10, 20, 30, 50\}$, the percentage of redundancy errors is very close to the result reported in Figure \ref{figure2}. When $k>30$, many generated sentences are completely incomprehensible. A larger $k$ will also reduce the generation efficiency. Therefore, we chose a sampling strategy of $k=30$, which is the trade-off between text quality and generation efficiency.

\section{Conclusions}
In this paper, we have presented TGEA, the first dataset with a variety of manual annotations on errors occurring texts generated by pretrained language models. For each erroneous text generated by a Chinese GPT-2 model, our crowdsourced annotators detect erroneous text spans with their associated text spans and provide error types defined in a two-level hierarchical taxonomy as well as rationales behind detected errors. We elaborate the 5 annotation stages for building TGEA with a strict annotation quality control protocol.  We also report baseline results of the 5 benchmark tasks on TGEA. The low results suggest that our dataset is a challenging testbed for future work on automatic detection of erroneous spans and types as well as producing error corrections and rationales for texts generated by PLMs. TGEA is featured with wide error type coverage, rich semantic annotation and functional diversity, which can not only be used for deep diagnostic analysis on the text generation capability of pretrained language models, but also facilitate and promote the research of automatic and interpretable error correction for PLM-generated texts.

\section*{Acknowledgments}

The present research was supported by Huawei. We would like to thank the anonymous reviewers for their insightful comments. We also want to thank MindSpore\footnote{2020. MindSpore. \href{https://www.mindspore.cn/}{https://www.mindspore.cn/}} for the partial suppoort of this work, which is a new deep learning computing framework. The corresponding author is Deyi Xiong (dyxiong@tju.edu.cn).

\normalem
\bibliography{acl2021}

\begin{thebibliography}{62}
\expandafter\ifx\csname natexlab\endcsname\relax\def\natexlab#1{#1}\fi

\bibitem[{Bernard and Han(2020)}]{bernard_mandarinograd_2020}
Timothée Bernard and Ting Han. 2020.
\newblock \href {https://www.aclweb.org/anthology/2020.lrec-1.3}
  {Mandarinograd: A chinese collection of winograd schemas}.
\newblock In \emph{Proceedings of the 12th Language Resources and Evaluation
  Conference}, pages 21--26. European Language Resources Association.

\bibitem[{Bhagavatula et~al.(2020)Bhagavatula, Le~Bras, Malaviya, Sakaguchi,
  Holtzman, Rashkin, Downey, Wen-tau Yi, and Choi}]{anli}
Chandra Bhagavatula, Ronan Le~Bras, Chaitanya Malaviya, Keisuke Sakaguchi, Ari
  Holtzman, Hannah Rashkin, Doug Downey, Scott Wen-tau Yi, and Yejin Choi.
  2020.
\newblock \href {https://arxiv.org/abs/1908.05739} {Abductive commonsense
  reasoning}.
\newblock \emph{International Conference on Learning Representations}.

\bibitem[{Bisk et~al.(2020)Bisk, Zellers, {LeBras}, Gao, and
  Choi}]{bisk_piqa_2020}
Yonatan Bisk, Rowan Zellers, Ronan {LeBras}, Jianfeng Gao, and Yejin Choi.
  2020.
\newblock \href {http://arxiv.org/abs/1911.11641} {{PIQA}: Reasoning about
  physical commonsense in natural language}.
\newblock In \emph{{AAAI}}, pages 7432--7439.

\bibitem[{Brown et~al.(2020)Brown, Mann, Ryder, Subbiah, Kaplan, Dhariwal,
  Neelakantan, Shyam, Sastry, Askell, Agarwal, Herbert-Voss, Krueger, Henighan,
  Child, Ramesh, Ziegler, Wu, Winter, Hesse, Chen, Sigler, Litwin, Gray, Chess,
  Clark, Berner, McCandlish, Radford, Sutskever, and Amodei}]{gpt3}
Tom Brown, Benjamin Mann, Nick Ryder, Melanie Subbiah, Jared~D Kaplan, Prafulla
  Dhariwal, Arvind Neelakantan, Pranav Shyam, Girish Sastry, Amanda Askell,
  Sandhini Agarwal, Ariel Herbert-Voss, Gretchen Krueger, Tom Henighan, Rewon
  Child, Aditya Ramesh, Daniel Ziegler, Jeffrey Wu, Clemens Winter, Chris
  Hesse, Mark Chen, Eric Sigler, Mateusz Litwin, Scott Gray, Benjamin Chess,
  Jack Clark, Christopher Berner, Sam McCandlish, Alec Radford, Ilya Sutskever,
  and Dario Amodei. 2020.
\newblock \href
  {https://proceedings.neurips.cc/paper/2020/file/1457c0d6bfcb4967418bfb8ac142f64a-Paper.pdf}
  {Language models are few-shot learners}.
\newblock In \emph{Advances in Neural Information Processing Systems},
  volume~33, pages 1876--1900. Curran Associates, Inc.

\bibitem[{Camburu et~al.(2018)Camburu, Rockt\"{a}schel, Lukasiewicz, and
  Blunsom}]{esnli}
Oana-Maria Camburu, Tim Rockt\"{a}schel, Thomas Lukasiewicz, and Phil Blunsom.
  2018.
\newblock \href
  {http://papers.nips.cc/paper/8163-e-snli-natural-language-inference-with-natural-language-explanations.pdf}
  {e-snli: Natural language inference with natural language explanations}.
\newblock In S.~Bengio, H.~Wallach, H.~Larochelle, K.~Grauman, N.~Cesa-Bianchi,
  and R.~Garnett, editors, \emph{Advances in Neural Information Processing
  Systems 31}, pages 9539--9549. Curran Associates, Inc.

\bibitem[{Cao et~al.(2020)Cao, Bi, Fang, and Tao}]{cao-etal-2020-pretrained}
Yu~Cao, Wei Bi, Meng Fang, and Dacheng Tao. 2020.
\newblock \href {https://doi.org/10.18653/v1/2020.findings-emnlp.81}
  {Pretrained language models for dialogue generation with multiple input
  sources}.
\newblock In \emph{Findings of the Association for Computational Linguistics:
  EMNLP 2020}, pages 909--917, Online. Association for Computational
  Linguistics.

\bibitem[{Chen et~al.(2019)Chen, Chu, and Gimpel}]{chen-etal-2019-evaluation}
Mingda Chen, Zewei Chu, and Kevin Gimpel. 2019.
\newblock \href {https://doi.org/10.18653/v1/D19-1060} {Evaluation benchmarks
  and learning criteria for discourse-aware sentence representations}.
\newblock In \emph{Proceedings of the 2019 Conference on Empirical Methods in
  Natural Language Processing and the 9th International Joint Conference on
  Natural Language Processing (EMNLP-IJCNLP)}, pages 649--662, Hong Kong,
  China. Association for Computational Linguistics.

\bibitem[{Cui et~al.(2020{\natexlab{a}})Cui, Wu, Liu, Zhang, and
  Zhou}]{cui-etal-2020-mutual}
Leyang Cui, Yu~Wu, Shujie Liu, Yue Zhang, and Ming Zhou. 2020{\natexlab{a}}.
\newblock \href {https://doi.org/10.18653/v1/2020.acl-main.130} {{M}u{T}ual: A
  dataset for multi-turn dialogue reasoning}.
\newblock In \emph{Proceedings of the 58th Annual Meeting of the Association
  for Computational Linguistics}, pages 1406--1416, Online. Association for
  Computational Linguistics.

\bibitem[{Cui et~al.(2020{\natexlab{b}})Cui, Che, Liu, Qin, Wang, and
  Hu}]{roberta-wwm-ext}
Yiming Cui, Wanxiang Che, Ting Liu, Bing Qin, Shijin Wang, and Guoping Hu.
  2020{\natexlab{b}}.
\newblock \href {https://www.aclweb.org/anthology/2020.findings-emnlp.58}
  {Revisiting pre-trained models for {C}hinese natural language processing}.
\newblock In \emph{Proceedings of the 2020 Conference on Empirical Methods in
  Natural Language Processing: Findings}, pages 657--668, Online. Association
  for Computational Linguistics.

\bibitem[{Dahlmeier and Ng(2012)}]{dahlmeier-ng-2012-better}
Daniel Dahlmeier and Hwee~Tou Ng. 2012.
\newblock \href {https://www.aclweb.org/anthology/N12-1067} {Better evaluation
  for grammatical error correction}.
\newblock In \emph{Proceedings of the 2012 Conference of the North {A}merican
  Chapter of the Association for Computational Linguistics: Human Language
  Technologies}, pages 568--572, Montr{\'e}al, Canada. Association for
  Computational Linguistics.

\bibitem[{Daudaravicius et~al.(2016)Daudaravicius, Banchs, Volodina, and
  Napoles}]{aesw}
Vidas Daudaravicius, Rafael~E. Banchs, Elena Volodina, and Courtney Napoles.
  2016.
\newblock \href {https://doi.org/10.18653/v1/W16-0506} {A report on the
  automatic evaluation of scientific writing shared task}.
\newblock In \emph{Proceedings of the 11th Workshop on Innovative Use of {NLP}
  for Building Educational Applications}, pages 53--62, San Diego, CA.
  Association for Computational Linguistics.

\bibitem[{Devlin et~al.(2019)Devlin, Chang, Lee, and
  Toutanova}]{devlin-etal-2019-bert}
Jacob Devlin, Ming-Wei Chang, Kenton Lee, and Kristina Toutanova. 2019.
\newblock \href {https://doi.org/10.18653/v1/N19-1423} {{BERT}: Pre-training of
  deep bidirectional transformers for language understanding}.
\newblock In \emph{Proceedings of the 2019 Conference of the North {A}merican
  Chapter of the Association for Computational Linguistics: Human Language
  Technologies, Volume 1 (Long and Short Papers)}, pages 4171--4186,
  Minneapolis, Minnesota. Association for Computational Linguistics.

\bibitem[{Fabbri et~al.(2019)Fabbri, Li, She, Li, and
  Radev}]{fabbri-etal-2019-multi}
Alexander Fabbri, Irene Li, Tianwei She, Suyi Li, and Dragomir Radev. 2019.
\newblock \href {https://doi.org/10.18653/v1/P19-1102} {Multi-news: A
  large-scale multi-document summarization dataset and abstractive hierarchical
  model}.
\newblock In \emph{Proceedings of the 57th Annual Meeting of the Association
  for Computational Linguistics}, pages 1074--1084, Florence, Italy.
  Association for Computational Linguistics.

\bibitem[{Flachs et~al.(2020)Flachs, Lacroix, Yannakoudakis, Rei, and
  S{\o}gaard}]{cweb}
Simon Flachs, Oph{\'e}lie Lacroix, Helen Yannakoudakis, Marek Rei, and Anders
  S{\o}gaard. 2020.
\newblock \href {https://doi.org/10.18653/v1/2020.emnlp-main.680} {Grammatical
  error correction in low error density domains: A new benchmark and analyses}.
\newblock In \emph{Proceedings of the 2020 Conference on Empirical Methods in
  Natural Language Processing (EMNLP)}, pages 8467--8478, Online. Association
  for Computational Linguistics.

\bibitem[{He et~al.(2022)He, Long, and Xiong}]{he-etal-2022-evaluating}
Jie He, Wanqiu Long, and Deyi Xiong. 2022.
\newblock \href {https://aclanthology.org/2022.codi-1.4/} {Evaluating discourse
  cohesion in pre-trained language models}.
\newblock In \emph{Proceedings of the 3rd Workshop on Computational Approaches
  to Discourse}, pages 28--34, Gyeongju, Republic of Korea and Online.
  International Conference on Computational Linguistics.

\bibitem[{He et~al.(2020)He, Liu, Gao, and Chen}]{he2020deberta}
Pengcheng He, Xiaodong Liu, Jianfeng Gao, and Weizhu Chen. 2020.
\newblock \href {http://arxiv.org/abs/2006.03654} {Deberta: Decoding-enhanced
  bert with disentangled attention}.

\bibitem[{Huang et~al.(2019)Huang, Le~Bras, Bhagavatula, and Choi}]{cosmosqa}
Lifu Huang, Ronan Le~Bras, Chandra Bhagavatula, and Yejin Choi. 2019.
\newblock \href {https://doi.org/10.18653/v1/D19-1243} {Cosmos {QA}: Machine
  reading comprehension with contextual commonsense reasoning}.
\newblock In \emph{Proceedings of the 2019 Conference on Empirical Methods in
  Natural Language Processing and the 9th International Joint Conference on
  Natural Language Processing (EMNLP-IJCNLP)}, pages 2391--2401, Hong Kong,
  China. Association for Computational Linguistics.

\bibitem[{Inoue et~al.(2020)Inoue, Stenetorp, and Inui}]{r4c}
Naoya Inoue, Pontus Stenetorp, and Kentaro Inui. 2020.
\newblock \href {https://doi.org/10.18653/v1/2020.acl-main.602} {{R}4{C}: A
  benchmark for evaluating {RC} systems to get the right answer for the right
  reason}.
\newblock In \emph{Proceedings of the 58th Annual Meeting of the Association
  for Computational Linguistics}, pages 6740--6750, Online. Association for
  Computational Linguistics.

\bibitem[{Iter et~al.(2020)Iter, Guu, Lansing, and
  Jurafsky}]{iter-etal-2020-pretraining}
Dan Iter, Kelvin Guu, Larry Lansing, and Dan Jurafsky. 2020.
\newblock \href {https://doi.org/10.18653/v1/2020.acl-main.439} {Pretraining
  with contrastive sentence objectives improves discourse performance of
  language models}.
\newblock In \emph{Proceedings of the 58th Annual Meeting of the Association
  for Computational Linguistics}, pages 4859--4870, Online. Association for
  Computational Linguistics.

\bibitem[{Jain et~al.(2020)Jain, van Zuylen, Hajishirzi, and
  Beltagy}]{jain-etal-2020-scirex}
Sarthak Jain, Madeleine van Zuylen, Hannaneh Hajishirzi, and Iz~Beltagy. 2020.
\newblock \href {https://doi.org/10.18653/v1/2020.acl-main.670} {{S}ci{REX}:
  {A} challenge dataset for document-level information extraction}.
\newblock In \emph{Proceedings of the 58th Annual Meeting of the Association
  for Computational Linguistics}, pages 7506--7516, Online. Association for
  Computational Linguistics.

\bibitem[{Jiao et~al.(2018)Jiao, Sun, and Sun}]{jiao2018LAC}
Zhenyu Jiao, Shuqi Sun, and Ke~Sun. 2018.
\newblock \href {https://arxiv.org/abs/1807.01882} {Chinese lexical analysis
  with deep bi-gru-crf network}.
\newblock \emph{arXiv preprint arXiv:1807.01882}.

\bibitem[{Kaneko et~al.(2020)Kaneko, Mita, Kiyono, Suzuki, and Inui}]{bert-gec}
Masahiro Kaneko, Masato Mita, Shun Kiyono, Jun Suzuki, and Kentaro Inui. 2020.
\newblock \href {https://doi.org/10.18653/v1/2020.acl-main.391}
  {Encoder-decoder models can benefit from pre-trained masked language models
  in grammatical error correction}.
\newblock In \emph{Proceedings of the 58th Annual Meeting of the Association
  for Computational Linguistics}, pages 4248--4254, Online. Association for
  Computational Linguistics.

\bibitem[{Levesque et~al.(2012)Levesque, Davis, and Morgenstern}]{wsc}
Hector~J. Levesque, Ernest Davis, and Leora Morgenstern. 2012.
\newblock \href
  {https://citeseerx.ist.psu.edu/viewdoc/download?doi=10.1.1.729.9814&rep=rep1&type=pdf}
  {The winograd schema challenge}.
\newblock In \emph{Proceedings of the Thirteenth International Conference on
  Principles of Knowledge Representation and Reasoning}, KR'12, page 552–561.
  AAAI Press.

\bibitem[{Lin(2004)}]{lin2004rouge}
Chin-Yew Lin. 2004.
\newblock \href {https://www.aclweb.org/anthology/W04-1013.pdf} {Rouge: A
  package for automatic evaluation of summaries}.
\newblock In \emph{Text summarization branches out}, pages 74--81.

\bibitem[{Liu and Lapata(2019)}]{liu-lapata-2019-text}
Yang Liu and Mirella Lapata. 2019.
\newblock \href {https://doi.org/10.18653/v1/D19-1387} {Text summarization with
  pretrained encoders}.
\newblock In \emph{Proceedings of the 2019 Conference on Empirical Methods in
  Natural Language Processing and the 9th International Joint Conference on
  Natural Language Processing (EMNLP-IJCNLP)}, pages 3730--3740, Hong Kong,
  China. Association for Computational Linguistics.

\bibitem[{Liu et~al.(2019)Liu, Ott, Goyal, Du, Joshi, Chen, Levy, Lewis,
  Zettlemoyer, and Stoyanov}]{roberta}
Yinhan Liu, Myle Ott, Naman Goyal, Jingfei Du, Mandar Joshi, Danqi Chen, Omer
  Levy, Mike Lewis, Luke Zettlemoyer, and Veselin Stoyanov. 2019.
\newblock \href {http://arxiv.org/abs/1907.11692} {Roberta: {A} robustly
  optimized {BERT} pretraining approach}.
\newblock \emph{CoRR}, abs/1907.11692.

\bibitem[{Long et~al.(2020{\natexlab{a}})Long, Cai, Reid, Webber, and
  Xiong}]{long-etal-2020-shallow}
Wanqiu Long, Xinyi Cai, James Reid, Bonnie Webber, and Deyi Xiong.
  2020{\natexlab{a}}.
\newblock \href {https://aclanthology.org/2020.lrec-1.129/} {Shallow discourse
  annotation for {C}hinese {TED} talks}.
\newblock In \emph{Proceedings of the Twelfth Language Resources and Evaluation
  Conference}, pages 1025--1032, Marseille, France. European Language Resources
  Association.

\bibitem[{Long et~al.(2024)Long, N, and Webber}]{long-etal-2024-multi}
Wanqiu Long, Siddharth N, and Bonnie Webber. 2024.
\newblock \href {https://doi.org/10.18653/v1/2024.findings-acl.500}
  {Multi-label classification for implicit discourse relation recognition}.
\newblock In \emph{Findings of the Association for Computational Linguistics:
  ACL 2024}, pages 8437--8451, Bangkok, Thailand. Association for Computational
  Linguistics.

\bibitem[{Long and Webber(2022)}]{long-webber-2022-facilitating}
Wanqiu Long and Bonnie Webber. 2022.
\newblock \href {https://doi.org/10.18653/v1/2022.emnlp-main.734} {Facilitating
  contrastive learning of discourse relational senses by exploiting the
  hierarchy of sense relations}.
\newblock In \emph{Proceedings of the 2022 Conference on Empirical Methods in
  Natural Language Processing}, pages 10704--10716, Abu Dhabi, United Arab
  Emirates. Association for Computational Linguistics.

\bibitem[{Long and
  Webber(2024)}]{long2024leveraginghierarchicalprototypesverbalizer}
Wanqiu Long and Bonnie Webber. 2024.
\newblock \href {http://arxiv.org/abs/2411.14880} {Leveraging hierarchical
  prototypes as the verbalizer for implicit discourse relation recognition}.

\bibitem[{Long et~al.(2020{\natexlab{b}})Long, Webber, and
  Xiong}]{long-etal-2020-ted}
Wanqiu Long, Bonnie Webber, and Deyi Xiong. 2020{\natexlab{b}}.
\newblock \href {https://doi.org/10.18653/v1/2020.emnlp-main.223} {{TED}-{CDB}:
  A large-scale {C}hinese discourse relation dataset on {TED} talks}.
\newblock In \emph{Proceedings of the 2020 Conference on Empirical Methods in
  Natural Language Processing (EMNLP)}, pages 2793--2803, Online. Association
  for Computational Linguistics.

\bibitem[{McHugh(2012)}]{kappa}
Mary McHugh. 2012.
\newblock \href {https://doi.org/10.11613/BM.2012.031} {Interrater reliability:
  The kappa statistic}.
\newblock \emph{Biochemia medica : časopis Hrvatskoga društva medicinskih
  biokemičara / HDMB}, 22:276--82.

\bibitem[{Napoles et~al.(2019)Napoles, Nadejde, and Tetreault}]{cmeg}
Courtney Napoles, Maria Nadejde, and Joel Tetreault. 2019.
\newblock \href {https://transacl.org/ojs/index.php/tacl/article/view/1677}
  {Enabling robust grammatical error correction in new domains: Datasets,
  metrics, and analyses}.
\newblock \emph{Transactions of the Association for Computational Linguistics},
  7(0):551--566.

\bibitem[{Napoles et~al.(2017)Napoles, Sakaguchi, and Tetreault}]{jfleg}
Courtney Napoles, Keisuke Sakaguchi, and Joel Tetreault. 2017.
\newblock \href {https://www.aclweb.org/anthology/E17-2037} {{JFLEG}: A fluency
  corpus and benchmark for grammatical error correction}.
\newblock In \emph{Proceedings of the 15th Conference of the {E}uropean Chapter
  of the Association for Computational Linguistics: Volume 2, Short Papers},
  pages 229--234, Valencia, Spain. Association for Computational Linguistics.

\bibitem[{Papineni et~al.(2002)Papineni, Roukos, Ward, and
  Zhu}]{papineni2002bleu}
Kishore Papineni, Salim Roukos, Todd Ward, and Wei-Jing Zhu. 2002.
\newblock \href {https://www.aclweb.org/anthology/P02-1040.pdf} {Bleu: a method
  for automatic evaluation of machine translation}.
\newblock In \emph{Proceedings of the 40th annual meeting of the Association
  for Computational Linguistics}, pages 311--318.

\bibitem[{Parikh et~al.(2020)Parikh, Wang, Gehrmann, Faruqui, Dhingra, Yang,
  and Das}]{parikh2020totto}
Ankur Parikh, Xuezhi Wang, Sebastian Gehrmann, Manaal Faruqui, Bhuwan Dhingra,
  Diyi Yang, and Dipanjan Das. 2020.
\newblock \href {https://doi.org/10.18653/v1/2020.emnlp-main.89} {{ToTTo}: A
  controlled table-to-text generation dataset}.
\newblock In \emph{Proceedings of the 2020 Conference on Empirical Methods in
  Natural Language Processing (EMNLP)}, pages 1173--1186, Online. Association
  for Computational Linguistics.

\bibitem[{Ponti et~al.(2020)Ponti, Glava{\v{s}}, Majewska, Liu, Vuli{\'c}, and
  Korhonen}]{xcopa}
Edoardo~Maria Ponti, Goran Glava{\v{s}}, Olga Majewska, Qianchu Liu, Ivan
  Vuli{\'c}, and Anna Korhonen. 2020.
\newblock \href {https://doi.org/10.18653/v1/2020.emnlp-main.185} {{XCOPA}: A
  multilingual dataset for causal commonsense reasoning}.
\newblock In \emph{Proceedings of the 2020 Conference on Empirical Methods in
  Natural Language Processing (EMNLP)}, pages 2362--2376, Online. Association
  for Computational Linguistics.

\bibitem[{Radford et~al.(2019)Radford, Wu, Child, Luan, Amodei, and
  Sutskever}]{gpt2}
Alec Radford, Jeff Wu, Rewon Child, David Luan, Dario Amodei, and Ilya
  Sutskever. 2019.
\newblock \href {http://www.persagen.com/files/misc/radford2019language.pdf}
  {Language models are unsupervised multitask learners}.

\bibitem[{Raffel et~al.(2020)Raffel, Shazeer, Roberts, Lee, Narang, Matena,
  Zhou, Li, and Liu}]{t5}
Colin Raffel, Noam Shazeer, Adam Roberts, Katherine Lee, Sharan Narang, Michael
  Matena, Yanqi Zhou, Wei Li, and Peter~J. Liu. 2020.
\newblock \href {http://jmlr.org/papers/v21/20-074.html} {Exploring the limits
  of transfer learning with a unified text-to-text transformer}.
\newblock \emph{Journal of Machine Learning Research}, 21(140):1--67.

\bibitem[{Roemmele et~al.(2011)Roemmele, Bejan, and
  Gordon}]{roemmele_choice_2011}
Melissa Roemmele, Cosmin~Adrian Bejan, and Andrew~S Gordon. 2011.
\newblock \href
  {https://ict.usc.edu/pubs/Choice%20of%20Plausible%20Alternatives-%20An%20Evaluation%20of%20Commonsense%20Causal%20Reasoning.pdf}
  {Choice of plausible alternatives: An evaluation of commonsense causal
  reasoning}.
\newblock In \emph{{AAAI} spring symposium: logical formalizations of
  commonsense reasoning}, pages 90--95.

\bibitem[{Sakaguchi et~al.(2020)Sakaguchi, Le~Bras, Bhagavatula, and
  Choi}]{sakaguchi_winogrande_2020}
Keisuke Sakaguchi, Ronan Le~Bras, Chandra Bhagavatula, and Yejin Choi. 2020.
\newblock \href {https://doi.org/10.1609/aaai.v34i05.6399} {{WinoGrande}: An
  adversarial winograd schema challenge at scale}.
\newblock In \emph{Proceedings of the {AAAI} Conference on Artificial
  Intelligence}, volume~34, pages 8732--8740.
\newblock Issue: 05.

\bibitem[{Sap et~al.(2019)Sap, Rashkin, Chen, Le~Bras, and Choi}]{siqa}
Maarten Sap, Hannah Rashkin, Derek Chen, Ronan Le~Bras, and Yejin Choi. 2019.
\newblock \href {https://doi.org/10.18653/v1/D19-1454} {Social {IQ}a:
  Commonsense reasoning about social interactions}.
\newblock In \emph{Proceedings of the 2019 Conference on Empirical Methods in
  Natural Language Processing and the 9th International Joint Conference on
  Natural Language Processing (EMNLP-IJCNLP)}, pages 4463--4473, Hong Kong,
  China. Association for Computational Linguistics.

\bibitem[{Sidorov et~al.(2020)Sidorov, Hu, Rohrbach, and
  Singh}]{image-to-caption}
Oleksii Sidorov, Ronghang Hu, Marcus Rohrbach, and Amanpreet Singh. 2020.
\newblock \href {http://arxiv.org/abs/2003.12462} {Textcaps: a dataset for
  image captioning with reading comprehension}.
\newblock \emph{CoRR}, abs/2003.12462.

\bibitem[{Talmor et~al.(2020)Talmor, Elazar, Goldberg, and Berant}]{olmpics}
Alon Talmor, Yanai Elazar, Yoav Goldberg, and Jonathan Berant. 2020.
\newblock \href {https://doi.org/10.1162/tacl\_a\_00342} {olmpics-on what
  language model pre-training captures}.
\newblock \emph{Transactions of the Association for Computational Linguistics},
  8:743--758.

\bibitem[{Talmor et~al.(2019)Talmor, Herzig, Lourie, and
  Berant}]{talmor_commonsenseqa_2019}
Alon Talmor, Jonathan Herzig, Nicholas Lourie, and Jonathan Berant. 2019.
\newblock \href {https://doi.org/10.18653/v1/N19-1421} {{CommonsenseQA}: A
  question answering challenge targeting commonsense knowledge}.
\newblock In \emph{Proceedings of the 2019 Conference of the North American
  Chapter of the Association for Computational Linguistics: Human Language
  Technologies, Volume 1 (Long and Short Papers)}, pages 4149--4158.
  Association for Computational Linguistics.

\bibitem[{Vougiouklis et~al.(2017)Vougiouklis, ElSahar, Kaffee, Gravier,
  Laforest, Hare, and Simperl}]{neu-wiki}
Pavlos Vougiouklis, Hady ElSahar, Lucie{-}Aim{\'{e}}e Kaffee, Christophe
  Gravier, Fr{\'{e}}d{\'{e}}rique Laforest, Jonathon~S. Hare, and Elena
  Simperl. 2017.
\newblock \href {http://arxiv.org/abs/1711.00155} {Neural wikipedian:
  Generating textual summaries from knowledge base triples}.
\newblock \emph{CoRR}, abs/1711.00155.

\bibitem[{Wang et~al.(2019)Wang, Pruksachatkun, Nangia, Singh, Michael, Hill,
  Levy, and Bowman}]{NEURIPS2019_4496bf24}
Alex Wang, Yada Pruksachatkun, Nikita Nangia, Amanpreet Singh, Julian Michael,
  Felix Hill, Omer Levy, and Samuel Bowman. 2019.
\newblock \href
  {https://proceedings.neurips.cc/paper/2019/file/4496bf24afe7fab6f046bf4923da8de6-Paper.pdf}
  {Superglue: A stickier benchmark for general-purpose language understanding
  systems}.
\newblock In \emph{Advances in Neural Information Processing Systems},
  volume~32, pages 3266--3280. Curran Associates, Inc.

\bibitem[{Wang et~al.(2018)Wang, Singh, Michael, Hill, Levy, and
  Bowman}]{wang-etal-2018-glue}
Alex Wang, Amanpreet Singh, Julian Michael, Felix Hill, Omer Levy, and Samuel
  Bowman. 2018.
\newblock \href {https://doi.org/10.18653/v1/W18-5446} {{GLUE}: A multi-task
  benchmark and analysis platform for natural language understanding}.
\newblock In \emph{Proceedings of the 2018 {EMNLP} Workshop {B}lackbox{NLP}:
  Analyzing and Interpreting Neural Networks for {NLP}}, pages 353--355,
  Brussels, Belgium. Association for Computational Linguistics.

\bibitem[{Wang et~al.(2020)Wang, Kurosawa, Katsumata, and
  Komachi}]{chinese-bert-gec}
Hongfei Wang, Michiki Kurosawa, Satoru Katsumata, and Mamoru Komachi. 2020.
\newblock \href {https://www.aclweb.org/anthology/2020.aacl-main.20} {{C}hinese
  grammatical correction using {BERT}-based pre-trained model}.
\newblock In \emph{Proceedings of the 1st Conference of the Asia-Pacific
  Chapter of the Association for Computational Linguistics and the 10th
  International Joint Conference on Natural Language Processing}, pages
  163--168, Suzhou, China. Association for Computational Linguistics.

\bibitem[{Warstadt et~al.(2020)Warstadt, Parrish, Liu, Mohananey, Peng, Wang,
  and Bowman}]{warstadt-etal-2020-blimp}
Alex Warstadt, Alicia Parrish, Haokun Liu, Anhad Mohananey, Wei Peng, Sheng-Fu
  Wang, and Samuel~R. Bowman. 2020.
\newblock \href {https://doi.org/10.1162/tacl_a_00321} {{BL}i{MP}: The
  benchmark of linguistic minimal pairs for {E}nglish}.
\newblock \emph{Transactions of the Association for Computational Linguistics},
  8:377--392.

\bibitem[{Weng et~al.(2020)Weng, Yu, Huang, Cheng, and
  Luo}]{Weng_Yu_Huang_Cheng_Luo_2020}
Rongxiang Weng, Heng Yu, Shujian Huang, Shanbo Cheng, and Weihua Luo. 2020.
\newblock \href {https://doi.org/10.1609/aaai.v34i05.6465} {Acquiring knowledge
  from pre-trained model to neural machine translation}.
\newblock \emph{Proceedings of the AAAI Conference on Artificial Intelligence},
  34(05):9266--9273.

\bibitem[{Wiseman et~al.(2017)Wiseman, Shieber, and
  Rush}]{wiseman-etal-2017-challenges}
Sam Wiseman, Stuart Shieber, and Alexander Rush. 2017.
\newblock \href {https://doi.org/10.18653/v1/D17-1239} {Challenges in
  data-to-document generation}.
\newblock In \emph{Proceedings of the 2017 Conference on Empirical Methods in
  Natural Language Processing}, pages 2253--2263, Copenhagen, Denmark.
  Association for Computational Linguistics.

\bibitem[{Xu et~al.(2020)Xu, Hu, Zhang, Li, Cao, Li, Xu, Sun, Yu, Yu, Tian,
  Dong, Liu, Shi, Cui, Li, Zeng, Wang, Xie, Li, Patterson, Tian, Zhang, Zhou,
  Liu, Zhao, Zhao, Yue, Zhang, Yang, Richardson, and Lan}]{cluewsc}
Liang Xu, Hai Hu, Xuanwei Zhang, Lu~Li, Chenjie Cao, Yudong Li, Yechen Xu, Kai
  Sun, Dian Yu, Cong Yu, Yin Tian, Qianqian Dong, Weitang Liu, Bo~Shi, Yiming
  Cui, Junyi Li, Jun Zeng, Rongzhao Wang, Weijian Xie, Yanting Li, Yina
  Patterson, Zuoyu Tian, Yiwen Zhang, He~Zhou, Shaoweihua Liu, Zhe Zhao, Qipeng
  Zhao, Cong Yue, Xinrui Zhang, Zhengliang Yang, Kyle Richardson, and Zhenzhong
  Lan. 2020.
\newblock \href {https://doi.org/10.18653/v1/2020.coling-main.419} {{CLUE}: A
  {C}hinese language understanding evaluation benchmark}.
\newblock In \emph{Proceedings of the 28th International Conference on
  Computational Linguistics}, pages 4762--4772, Barcelona, Spain (Online).
  International Committee on Computational Linguistics.

\bibitem[{Yannakoudakis et~al.(2011)Yannakoudakis, Briscoe, and Medlock}]{fce}
Helen Yannakoudakis, Ted Briscoe, and Ben Medlock. 2011.
\newblock \href {https://www.aclweb.org/anthology/P11-1019} {A new dataset and
  method for automatically grading {ESOL} texts}.
\newblock In \emph{Proceedings of the 49th Annual Meeting of the Association
  for Computational Linguistics: Human Language Technologies}, pages 180--189,
  Portland, Oregon, USA. Association for Computational Linguistics.

\bibitem[{Zellers et~al.(2019{\natexlab{a}})Zellers, Bisk, Farhadi, and
  Choi}]{zellers2019recognition}
Rowan Zellers, Yonatan Bisk, Ali Farhadi, and Yejin Choi. 2019{\natexlab{a}}.
\newblock \href
  {http://openaccess.thecvf.com/content_CVPR_2019/html/Zellers_From_Recognition_to_Cognition_Visual_Commonsense_Reasoning_CVPR_2019_paper.html}
  {From recognition to cognition: Visual commonsense reasoning}.
\newblock In \emph{Proceedings of the IEEE/CVF Conference on Computer Vision
  and Pattern Recognition}, pages 6720--6731.

\bibitem[{Zellers et~al.(2019{\natexlab{b}})Zellers, Holtzman, Bisk, Farhadi,
  and Choi}]{hellaswag}
Rowan Zellers, Ari Holtzman, Yonatan Bisk, Ali Farhadi, and Yejin Choi.
  2019{\natexlab{b}}.
\newblock \href {https://doi.org/10.18653/v1/P19-1472} {{H}ella{S}wag: Can a
  machine really finish your sentence?}
\newblock In \emph{Proceedings of the 57th Annual Meeting of the Association
  for Computational Linguistics}, pages 4791--4800, Florence, Italy.
  Association for Computational Linguistics.

\bibitem[{Zhang et~al.(2020{\natexlab{a}})Zhang, Zhao, and Song}]{winowhy}
Hongming Zhang, Xinran Zhao, and Yangqiu Song. 2020{\natexlab{a}}.
\newblock \href {https://doi.org/10.18653/v1/2020.acl-main.508} {{W}ino{W}hy: A
  deep diagnosis of essential commonsense knowledge for answering {W}inograd
  schema challenge}.
\newblock In \emph{Proceedings of the 58th Annual Meeting of the Association
  for Computational Linguistics}, pages 5736--5745, Online. Association for
  Computational Linguistics.

\bibitem[{Zhang et~al.(2020{\natexlab{b}})Zhang, Lyu, and
  Callison-Burch}]{zhang-etal-2020-reasoning}
Li~Zhang, Qing Lyu, and Chris Callison-Burch. 2020{\natexlab{b}}.
\newblock \href {https://doi.org/10.18653/v1/2020.emnlp-main.374} {Reasoning
  about goals, steps, and temporal ordering with {W}iki{H}ow}.
\newblock In \emph{Proceedings of the 2020 Conference on Empirical Methods in
  Natural Language Processing (EMNLP)}, pages 4630--4639, Online. Association
  for Computational Linguistics.

\bibitem[{Zhang et~al.(2020{\natexlab{c}})Zhang, Kishore, Wu, Weinberger, and
  Artzi}]{bert-score}
Tianyi Zhang, Varsha Kishore, Felix Wu, Kilian~Q. Weinberger, and Yoav Artzi.
  2020{\natexlab{c}}.
\newblock \href {https://openreview.net/forum?id=SkeHuCVFDr} {Bertscore:
  Evaluating text generation with bert}.
\newblock In \emph{International Conference on Learning Representations}.

\bibitem[{Zhang et~al.(2021)Zhang, Yang, and Zhao}]{zhang2021retro}
Zhuosheng Zhang, Junjie Yang, and Hai Zhao. 2021.
\newblock \href {https://arxiv.org/abs/2001.09694} {Retrospective reader for
  machine reading comprehension}.
\newblock In \emph{The Thirty-Fifth AAAI Conference on Artificial Intelligence
  (AAAI-21)}.

\bibitem[{Zhao et~al.(2018)Zhao, Jiang, Sun, and Wan}]{cgec}
Yuanyuan Zhao, Nan Jiang, Weiwei Sun, and Xiaojun Wan. 2018.
\newblock \href {https://doi.org/10.1007/978-3-319-99501-4_41} {\emph{Overview
  of the NLPCC 2018 Shared Task: Grammatical Error Correction: 7th CCF
  International Conference, NLPCC 2018, Hohhot, China, August 26–30, 2018,
  Proceedings, Part II}}, pages 439--445.

\bibitem[{Zhou et~al.(2020)Zhou, Zhang, Cui, and
  Huang}]{DBLP:conf/aaai/ZhouZCH20}
Xuhui Zhou, Yue Zhang, Leyang Cui, and Dandan Huang. 2020.
\newblock \href {https://aaai.org/ojs/index.php/AAAI/article/view/6523}
  {Evaluating commonsense in pre-trained language models}.
\newblock In \emph{The Thirty-Fourth {AAAI} Conference on Artificial
  Intelligence, {AAAI} 2020, The Thirty-Second Innovative Applications of
  Artificial Intelligence Conference, {IAAI} 2020, The Tenth {AAAI} Symposium
  on Educational Advances in Artificial Intelligence, {EAAI} 2020, New York,
  NY, USA, February 7-12, 2020}, pages 9733--9740. {AAAI} Press.

\end{thebibliography}
\bibliographystyle{acl_natbib}
\appendix
\setcounter{table}{0} 
\section{Appendix}
\subsection{NEZHA-Gen Hyperparameters}
Table  \ref{gen} show  the  configuration of  the  generative model (NEZHA-Gen).
\begin{table}[!h]

\small
    \centering
    \begin{tabular}{c|c}
    \bottomrule
         Model&NEZHA-Gen \\
         \hline
         hidden\_size&768\\
         num\_hidden\_layers&12\\
         num\_attention\_heads&12\\
         intermediate\_size&3072\\
         hidden\_act&gelu\\
         hidden\_dropout\_prob&0.1\\
         attention\_probs\_dropout\_prob&0.1\\
         max\_position\_embeddings&512\\
         type\_vocab\_size&16\\
         initializer\_range&0.02\\
    \toprule
    \end{tabular}
    \caption{Configuration of NEZHA-Gen.}
    \label{gen}
\end{table}

\subsection{Training Setting}
Table \ref{tab:etd_para}, \ref{tab:easd_para}, \ref{tab:etc_para}, \ref{tab:gec_para}, \ref{tab:rg_para} show the training settings of the baseline models for each task. In these tables, $\rm ALBERT_{zh}$, $\rm BERT_{zh}$, $\rm RoBERTa_{zh}$ represent ALBERT-chinese, RoBerta-wwm-ext and RoBerta-wwm-ext respectively.
\begin{table}[h]

\small
    \centering
    \begin{tabular}{c|ccc}
    \bottomrule
         Model&$\rm ALBERT_{zh}$&$\rm BERT_{zh}$& $\rm RoBERTa_{zh}$ \\
         \hline 
         Model size&large&base&large\\
         Learning rate&\multicolumn{3}{c}{$2\times 10^{-5}$}\\
         Batch size&\multicolumn{3}{c}{8}\\
         Optimizer&\multicolumn{3}{c}{Adam}\\
         Adam $\beta_1$&\multicolumn{3}{c}{$0.9$}\\
         Adam $\beta_2$&\multicolumn{3}{c}{$0.98$}\\
         Adam $\epsilon$&\multicolumn{3}{c}{$1 \times 10^{-8}$}\\
         Max epochs&\multicolumn{3}{c}{50}\\
         Loss function&\multicolumn{3}{c}{cross-entropy}\\
         Dropout&\multicolumn{3}{c}{0.1}\\
    \toprule
    \end{tabular}
    \caption{Training details for the Erroneous Text Detection task.}
    \label{tab:etd_para}
\end{table}
\begin{table}[h]

    \small
    \begin{tabular}{c|ccc}
    \bottomrule
         Model&$\rm ALBERT_{zh}$&$\rm BERT_{zh}$& $\rm RoBERTa_{zh}$ \\
         \hline 
         Model size&base&base&base\\
         Learning rate&\multicolumn{3}{c}{$2\times 10^{-5}$}\\
         Batch size&\multicolumn{3}{c}{32}\\
         Optimizer&\multicolumn{3}{c}{Adam}\\
         Adam $\beta_1$&\multicolumn{3}{c}{$0.9$}\\
         Adam $\beta_2$&\multicolumn{3}{c}{$0.999$}\\
         Adam $\epsilon$&\multicolumn{3}{c}{$1 \times 10^{-6}$}\\
         Max epochs&\multicolumn{3}{c}{5}\\
         Loss function&\multicolumn{3}{c}{cross-entropy}\\
         Dropout&\multicolumn{3}{c}{0.1}\\
    \toprule
    \end{tabular}
    \caption{Training details for the Erroneous and Associated Span Detection task.}
    \label{tab:easd_para}
\end{table}
\begin{table}[thp]

\scriptsize
    \centering
    \begin{tabular}{c|ccc}
    \bottomrule
         Model&$\rm ALBERT_{zh}$&$\rm BERT_{zh}$& $\rm RoBERTa_{zh}$ \\
         \hline 
         Model size&large&base&large\\
         Learning rate&\multicolumn{3}{c}{$2\times10^{-5}$}\\
         Batch size&\multicolumn{3}{c}{8}\\
         Optimizer&\multicolumn{3}{c}{Adam}\\
         Adam $\beta_1$&\multicolumn{3}{c}{$0.9$}\\
         Adam $\beta_2$&\multicolumn{3}{c}{$0.98$}\\
         Adam $\epsilon$&\multicolumn{3}{c}{$1 \times 10^{-8}$}\\
         Max epochs&\multicolumn{3}{c}{50}\\
         Loss function&\multicolumn{3}{c}{cross-entropy}\\
         Dropout&\multicolumn{3}{c}{0.1}\\
    \toprule
    \end{tabular}
    \caption{Training details for the Error Type Classification task.}
    \label{tab:etc_para}
\end{table}
\begin{table}[!h]
\scriptsize
    \centering
    \begin{tabular}{c|cc}
    \bottomrule
         &$\rm BERT_{zh}$\_GEC& $\rm RoBERTa_{zh}$\_GEC \\
         \hline 
         Model&BERT-wwm-ext&RoBERTa-wwm-ext-large\\
         Architecture&\multicolumn{2}{c}{Transformer (big)}\\
         Learning rate&\multicolumn{2}{c}{$3\times10^{-5}$}\\
         Batch size&\multicolumn{2}{c}{16}\\
         Optimizer&\multicolumn{2}{c}{Adam}\\
         Adam $\beta_1$&\multicolumn{2}{c}{$0.9$}\\
         Adam $\beta_2$&\multicolumn{2}{c}{$0.98$}\\
         Adam $\epsilon$&\multicolumn{2}{c}{$1 \times 10^{-8}$}\\
         Max epochs&\multicolumn{2}{c}{50}\\
         Loss function&\multicolumn{2}{c}{label smoothed cross-entropy ($\epsilon_{ls} = 0.1)$}\\
         Dropout&\multicolumn{2}{c}{0.3}\\
    \toprule
    \end{tabular}
    \caption{Training details for the Error Correction  task.}
    \label{tab:gec_para}
\end{table}
\begin{table}[!h]

\small
    \centering
    \begin{tabular}{c|c}
    \bottomrule
         Model&NEZHA-Gen \\
         \hline
         Learning rate&$5 \times 10^{-5}$\\
         Batch size&4\\
         Optimizer&Adam\\
         Adam $\beta_1$&$0.9$\\
         Adam $\beta_2$&$0.999$\\
         Adam $\epsilon$&$1 \times 10^{-6}$\\
         Max epochs&3\\
         Dropout&0.1\\
    \toprule
    \end{tabular}
    \caption{Training details for the Rationale Generation task.}
    \label{tab:rg_para}
\end{table}

\subsection{Examples of Level-2 Error Types}
Table \ref{tab:examples} shows examples of level-2 error types in TGEA.
\begin{table*}[tbp]
\tiny
    \centering
    \begin{tabular}{|c|c|p{40em}|}
\bottomrule
Level-1 Error Type&Level-2 Error Type&Example \\
\hline
\multirow{10}{*}{\makecell{Inappropriate\\Combination}}&\multirow{2}{*}{\makecell{Subject-Predicate}}&目前,该市的\uwave{小说} [话剧]《我是党员、我的团员》、《我是小老头》、《小小老师》、《小小一个农家娃》\uline{正在上演}。\\
&& At present, the city’s \uwave{novels} [drama] \textit{I am a Party member and This is My League Member}, \textit{Little Old Man Like Me}, \textit{Little Teacher}, \textit{A Little Farm Boy} \uline{are on stage}. \\  
\cline{2-3}
&\multirow{2}{*}{\makecell{Predicate-Object}}&由我主持，我要带大家去\uline{感受}一下大赛主题设置的\uwave{感受} [氛围]。\\
&& As a host, I will take you to \uline{experience} the \uwave{feel} [atmosphere] shown from the theme of the competition.  \\  
\cline{2-3}
&\multirow{2}{*}{\makecell{Subject-Object}}&女足的\uwave{队员} [任务] 就是\uline{一个球}，能够把球踢好，就是她们最大的资本。\\
&&The \uwave{players} [task] of women’s football team is \uline{a ball}, and playing the ball well is their biggest capitals.\\
\cline{2-3}
&\multirow{2}{*}{\makecell{Modifier}}&另一方面，煤炭企业面临着\uline{煤矿安全}的\uwave{矛盾} [问题]。\\
&&On the other hand, coal enterprises are facing the \uwave{contradiction} [problem]  of \uline{coal mine safety}.\\
\cline{2-3} 
&\multirow{2}{*}{\makecell{Function Word}}&因此，我\uwave{对} [因为]自身的过错\uline{作出}了自己应当承担的责任。\\
&&Therefore, \uwave{to} [because of] my own fault, I \uline{took} my own responsibility.\\
\hline 
\multirow{10}{*}{Misssing}&\multirow{2}{*}{\makecell{Subject}}&当他回到车间时，\uwave{~~~}[车间]已经\uline{有了}明显的变化。\\
&&When he returned to the workshop, \uwave{~~~}[the place] \uline{had been} a marked change\\
\cline{2-3} 
&\multirow{2}{*}{\makecell{Predicate}}&这时候我们一开始就有机会扳平比分，但是\uline{我们没有}\uwave{~~~}[抓住]机会。\\
&&We had a chance to equalise at the beginning, but \uline{we didn't} \uwave{~~~} [caught] chance. \\
\cline{2-3}  
&\multirow{2}{*}{\makecell{Object}}&一、坚持解放思想,转变观念,\uline{推进}社会主义物质文明和精神\uwave{~~~}[文明]。\\
&&1. Persisting in emancipating the mind,  changing ideas and  \uline{promoting} socialist material civilization and spiritual \uwave{~~~} [civilization]. \\
\cline{2-3}  
&\multirow{2}{*}{\makecell{Modifier}}&在国内成立水牛研究中心，有利于增强\uwave{~~~}[水牛对]自然条件和人工环境的\uline{适应能力}。\\
&&The establishment of Buffalo Research Center in China is conducive to enhance \uline{the adaptability} \uwave{~~~} [of buffalo] to natural conditions and artificial environment.\\
\cline{2-3}  
&\multirow{2}{*}{\makecell{Function Word}}&他的儿子\uwave{~~~}[在]上一届\uline{奥运会}夺得冠军，并且获得当年世界锦标杯赛金牌。\\
&&His son won champion \uwave{~~~}[in] the last \uline{Olympic Games} and won the gold medal in the World Championship Cup that year.  \\
\hline
\multirow{10}{*}{Redundancy}&\multirow{2}{*}{\makecell{Subject}}&但\uline{一些外资银行}\uwave{，尤其是外资银行}[]，对我国民营经济的发展还有不少误解或偏见。\\
&&However, \uline{some foreign banks}\uwave{, especially foreign banks}[], still have many misunderstandings or prejudices about the development of China's private economy. \\
\cline{2-3}   
&\multirow{2}{*}{\makecell{Predicate}}&这也是所有\uwave{关心}[]\uline{关心}孩子成长的人的共同心声。\\
&&This is also the common voice of all those who \uwave{care about}[] \uline{care about} children's growth\\
\cline{2-3}   
&\multirow{2}{*}{\makecell{Object}}&同时，学校也开展丰富多彩、有益于学生的\uline{社会实践}活动\uwave{、社会实践}[]，丰富他们的课余生活。\\
&&At the same time, the school also carries out colorful and beneficial \uline{social practice} activities\uwave{, social practice}[] to enrich their after-school life. \\
\cline{2-3}   
&\multirow{2}{*}{\makecell{Modifier}}&它们的皮毛很有光泽,可以用肉眼\uwave{很难}[]\uline{看出来}。\\
&&Their fur is so shiny that we can \uline{see} with naked eyes \uwave{hardly}[]. \\
\cline{2-3}
&\multirow{2}{*}{\makecell{Function Word}}&他是被迫进入位于市中心的一个警察局的，\uwave{随后}[]\uline{他被带到警察局}，并遭到了手铐和警犬的威吓。\\
&&He was forced into a police station in the center of the city,  \uwave{then}[] \uline{he was taken to the police station}, where he was intimidated by handcuffs and police dogs.\\
\hline
\multirow{2}{*}{\makecell{Discourse\\Error}}&\multirow{2}{*}{\makecell{Coreference}}&在\uline{婚姻}变得更为不好的时候，对她来说这是痛苦的。但是当\uwave{她}[它]发生变化时，她必须做出调整。\\
&&It was painful for her when \uline{the marriage} got worse. But when \uwave{she} [it] changed, she had to adjust. \\
\hline  
\multirow{16}{*}{\makecell{Commonsense\\Error}}&\multirow{2}{*}{\makecell{Space}}&他说,\uline{中美}两国是\uwave{近邻} [朋友],关系很好,中美合作富有创造性。\\
&&He said that \uline{China and the United States} are close \uwave{neighbors} [friends] with good relations and creative cooperation.\\
\cline{2-3}   
&\multirow{2}{*}{\makecell{Time}}&\uwave{国庆} [元旦]假期期间，各大汽车经销商将会以怎么样的姿态迎接\uline{新的一年}？\\
&&During the \uwave{National Day} [New Year's Day] holiday, how will major auto dealers greet \uline{the new year}? \\
\cline{2-3}   
&\multirow{2}{*}{\makecell{Number}}&而在4月份，\uline{中国石化、招商银行、万科、上海汽车、g长安和g天威}成为了最活跃的\uwave{5} [6]只股票。\\
&&In April, \uline{Sinopec, China Merchants Bank, Vanke, SAIC, G Changan and G Tianwei} became the most active \uwave{5} [6] stocks. \\
\cline{2-3}   
&\multirow{2}{*}{\makecell{Motivation}}&近日，李老的胃疼难忍，\uline{为治疗病情}已连续\uwave{工作} [休息]两天了，而且病情非常严重，他一躺就是几天。\\
&&Recently, Lao Li’s stomach ache is unbearable. He has been \uwave{working} [resting] for two consecutive days \uline{to treat his illness}, and his illness is very serious. He has been lying down for several days. \\
\cline{2-3}   
&\multirow{2}{*}{\makecell{Emotional Reactions}}&对于学校\uline{为了保障}广大师生员工的安全，采取这些措施，我们深感\uwave{遗憾}[欣慰]。\\
&&We are very \uwave{sorry} [pleased] that the school has taken these measures \uline{to ensure} the safety of students, teachers, and other staff.\\
\cline{2-3}   
&\multirow{2}{*}{\makecell{Causation }}&据悉，由于身价\uwave{低廉}[高昂]，子淇在国内是很少有人请得到的\uline{大牌艺人}之一。\\
&&It is reported that Ziqi is one of the few \uline{famous artists} that are difficult to invite in China because of his \uwave{low} [high] value.\\
\cline{2-3}   
&\multirow{2}{*}{\makecell{Taxonomy}}&\uwave{酱} [花生] 油是\uline{植物油}中的一种，食用后可以对皮肤有非常好的润泽效果。\\
&&\uwave{Soy sauce} [Peanut Oil] is a kind of \uline{vegetable oil}, which has a very good moisturizing effect on the skin after eating. \\
\cline{2-3}   
&\multirow{2}{*}{\makecell{ Behaviors}}&一位\uline{中国}官员表示：我们将在近期和俄罗斯\uwave{、中国} [法国] 等国合作进一步推广这一系列行动，以此来缓解人们对恐怖主义威胁的忧虑。\\
&&In the near future, we will work with Russia, \uwave{China} [France] and other countries to further promote this series of actions to ease people's concerns about the threat of terrorism, a \uline{Chinese} official said.\\
\toprule
    \end{tabular}
    \caption{Examples of level-2 error types in TGEA.  \uwave{Underwaved words} are erroneous words while \uline{underlined words} are associated words. Words in "[]" are corrections to erroneous words.}
    \label{tab:examples}
 \end{table*}
\end{CJK}
\end{document}